\documentclass{llncs}
\usepackage{lineno,hyperref}
\usepackage{color}
\usepackage{amsfonts}
\usepackage{amsmath}
\usepackage{amssymb}
\usepackage{bm}
\usepackage{mathabx} 
\usepackage{multirow}
\usepackage{subcaption}
\usepackage{graphicx}
\usepackage{rotating}
\usepackage{listings}
\usepackage{tikz, pgf}
\usepackage{tikz-3dplot}
\usepackage[linesnumbered, ruled]{algorithm2e}

\makeatletter
\def\@xfootnote[#1]{%
  \protected@xdef\@thefnmark{#1}%
  \@footnotemark\@footnotetext}
\makeatother
\modulolinenumbers[5]

\begin{document}

\title{A Gaussian Process Upsampling Model for Improvements in Optical Character Recognition}
\author{Steven I. Reeves\inst{1,2} \and Dongwook Lee\inst{1} \and Anurag Singh\inst{2} \and Kunal Verma\inst{2} }
\institute{University of California, Santa Cruz \email{sireeves@ucsc.edu} \and AppZen Inc.}

\maketitle

\begin{abstract}
Optical Character Recognition and extraction is a key tool in the automatic evaluation of documents in a financial context. However,
the image data provided to automated systems can have unreliable quality, and can be inherently low-resolution or downsampled and compressed 
by a transmitting program. In this paper, we illustrate the efficacy of a Gaussian Process upsampling model for the purposes of improving OCR and 
extraction through upsampling low resolution documents. 
\end{abstract}




\section{Introduction}
An important problem in computer vision is the retrieval of textual information from images of documents. This is especially useful 
for search engines, accessibility tools for the visually impaired, and for processing of financial documents. For these purposes, 
optical character recognition (OCR) engines have been constructed. The popular open-source OCR framework Tesseract is used in 
this study. Optical character recognition frameworks, in general, are only as good as the document image that is supplied to them. 
In many cases, the resolution of the 
document image plays a role in how well the characters are extracted.
 In order to better upsample low resolution document images, 
 a new Gaussian Process Modeling upsampling algorithm is constructed and presented in this paper. 

This manuscript is organized in the following way. To begin, 
OCR is introduced along with the state-of-the-art OCR extraction software Tesseract. Next, the Gaussian Process based 
upsampling method is discussed, along with a brief study on the choice of covariance kernels and the use of a maximum likelihood estimate
for the mean. Finally the algorithm is tested against the baseline bicubic upsampling technique by examining the produced OCR accuracy resulting 
from these upsampled images. 

\section{Optical Character Recognition}
 Optical Character Recognition is the conversion of pixel represented words and characters within images into machine-encoded text. 
As previously mentioned, the OCR framework Tesseract~\cite{tesseract} is used to extract text in the document images used in this manuscript. 
Tesseract was originally formulated by HP research between 1984 and 1994. Since then it has changed hands and now is an open-source 
software package managed by Google~\cite{tess2} -- under the Apache 2.0 License. 
We use Tesseract 4.1.1, which generates text based utilizing  a Long-Short Term
Memory (LSTM) network. Tesseract ingests single-channel images and generates feature-maps based on these images. Then these feature maps 
are embedded into an input for the LSTM~\cite{tesseract,tess2}. 

\section{The Gaussian Process Algorithm}
\label{sec:GP25}
This interpolation method has taken inspiration from a new interpolation method for computation fluid dynamics proposed in \cite{GPAMR}. 
The authors used a windowed GP method to upsample simulation data from coarse to fine computational meshes.
For our application we define text in single-channel document images by pixels with low intensity values (close to 0 or black), 
surrounded by pixels of high intensity (closer to 255 or white in an 8 bit context). 
Specifically, pixel values are low in the interior of a character, and pixel values are 
comparatively high outside of characters. Because of this specific structure, the type of GP modeling will change. Instead of 
modeling the raw values, the deviation from a mean intensity will be modeled. This allows the upsampling algorithm to better maintain these 
intensities in the presence of characters. This structure is discussed in more detail in Subsections~\ref{subsec:MLE} and \ref{subsec:OCRalgo}. 
Mathematically, we define the upsampling operator to be 
\begin{equation}
f_* = f_0 + \mathbf{k}_*^T\mathbf{K}^{-1}\left(\mathbf{f} - \bar{\mathbf{f}}\right)
\label{eq:mean}
\end{equation}
which follows the formula for the posterior mean~\cite{Rasmussen2005}. In Equation~(\ref{eq:mean}), $\mathbf{k}_*$ is a vector of covariances
between the sample pixel locations and the location of the pixel we wish to interpolate. Furthermore $\mathbf{K}$ is a matrix of pairwise 
covariances between sample points. The term $f_0$ is the estimate for the prior mean pixel intensity over the sample, and $\bar{\mathbf{f}} = 
f_0\mathbf{1}$, calculation of these terms is found in Subsection~\ref{subsec:MLE}. In Subsection~\ref{subsec:Cov}, we discuss the choice of 
covariance kernel to generate $\mathbf{k}_*$ and $\mathbf{K}$.

\subsection{Choice of Covariance Kernel}
\label{subsec:Cov}
The commonly used squared exponential kernel~\cite{Rasmussen2005} is often used when the underlying function is continuous and is the 
de-facto covariance function when building a Gaussian Process. 
Image data on the other hand, is inherently discontinuous and is comprised of 8 or 16 bit integers.
So instead of the SE kernel, a member of Mat\'{e}rn family of kernels is used.
In the Mat\'{e}rn  family of covariance functions, there are 3 hyper-parameters that dictate their character -- as indicated in 
Equation~(\ref{eq:matern}).
\begin{equation}
K_{mat}(\mathbf{x}, \mathbf{y}) = \Sigma^2 \frac{2^{1-\nu}}{\Gamma(\nu)}\left(\sqrt{2\nu}
\frac{||\mathbf{x} - \mathbf{y}||}{\ell}\right)^\nu K_{\nu} \left(\sqrt{2\nu}
\frac{||\mathbf{x} - \mathbf{y}||}{\ell}\right)
\label{eq:matern}
\end{equation}

For the Mat\'{e}rn kernels, there are three hyper-parameters $\Sigma$, $\ell$, and $\nu$.
The hyper-parameter $\Sigma$ related to the output variance function, and is widely used for uncertainty quantification. 
The term $\ell$ is the inherent length scale of covariance in for the underlying function space. 
 The hyper-parameter $\nu$ on the other hand, relates the level of "continuousness" of the functions that are 
sampled. The function $K_\nu$ is the modified Bessel function of the second kind of order $\nu$.  
The Mat\'{e}rn family of covariance functions give continuity properties ranging infinitely differentiable 
functions, as produced by the SE kernel, and nowhere differentiable -- such as those generated by the Ornstein-Uhlenbeck covariance kernel.

Consideration of the input and output datatypes of the Gaussian Process are key when choosing or building a covariance function.
The datatype for this application are document images, which contain sharp contrasts that are handled better by a low $\nu$ Mat\'{e}rn kernel.
 The Mat\'{e}rn kernel with $\nu = 3/2$ is used in this algorithm. For this specific value, Equation~(\ref{eq:matern}) can be simplified. 
By setting $\nu = 3/2$, 
\begin{equation}
K_{3/2}(\mathbf{x}, \mathbf{y}) = \Sigma^2\left(1 + \sqrt{3}\frac{||\mathbf{x} - \mathbf{y}||}{\ell} \right)
 \exp\left(-\sqrt{3}\frac{||\mathbf{x} - \mathbf{y}||}{\ell}\right).
\label{eq:matern3/2}
\end{equation}
We choose $\Sigma = 1$ as the uncertainty portion of GP modeling will not be used for this application.

\begin{figure}
\begin{center}
\begin{subfigure}{0.45\textwidth}
\includegraphics[width=\textwidth]{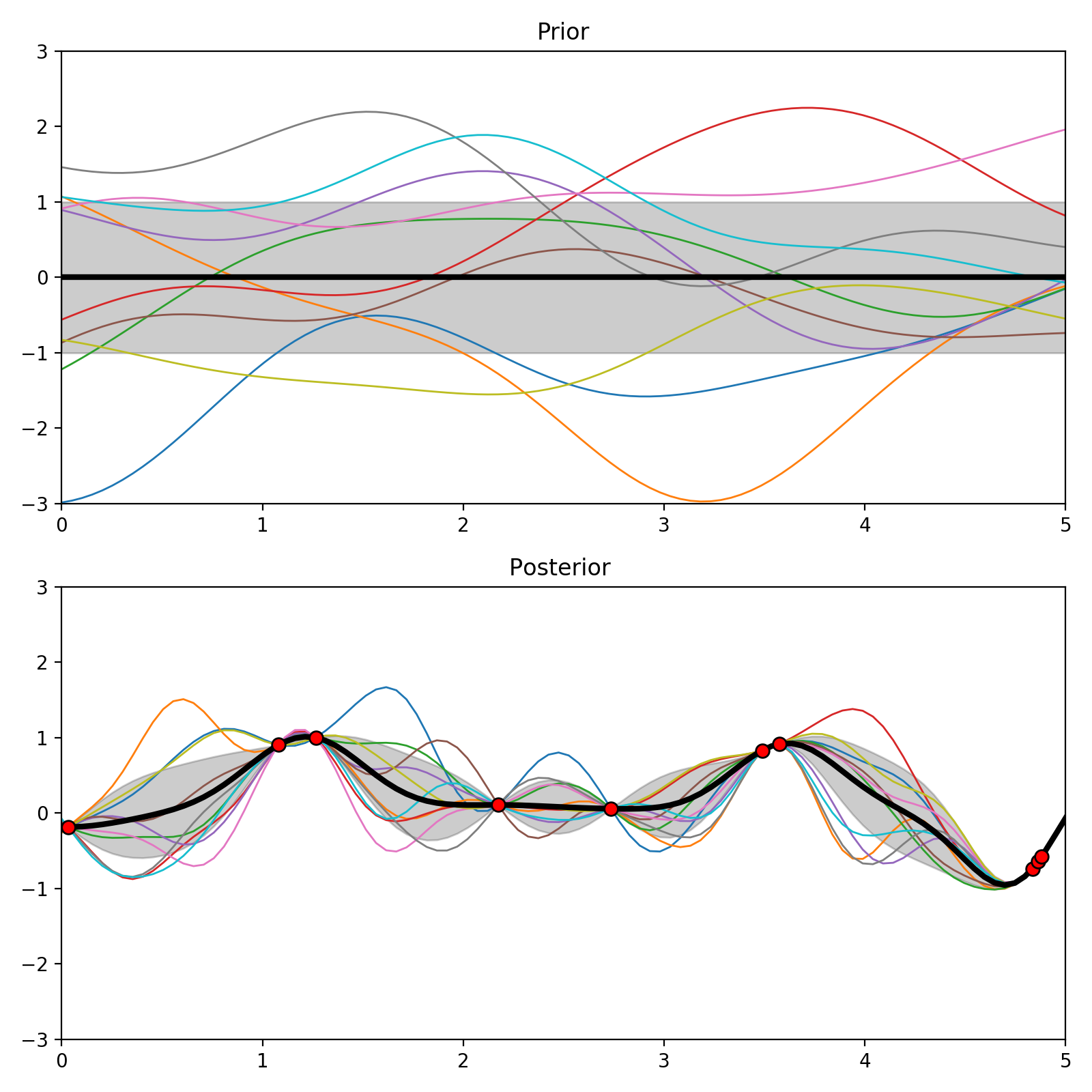}
\caption{\label{fig:sqrexp_samp} Top: Prior Mean functions.
Bottom: Posterior mean functions.}
\end{subfigure}%
\hfill
\begin{subfigure}{0.45\textwidth}
\includegraphics[width=\textwidth]{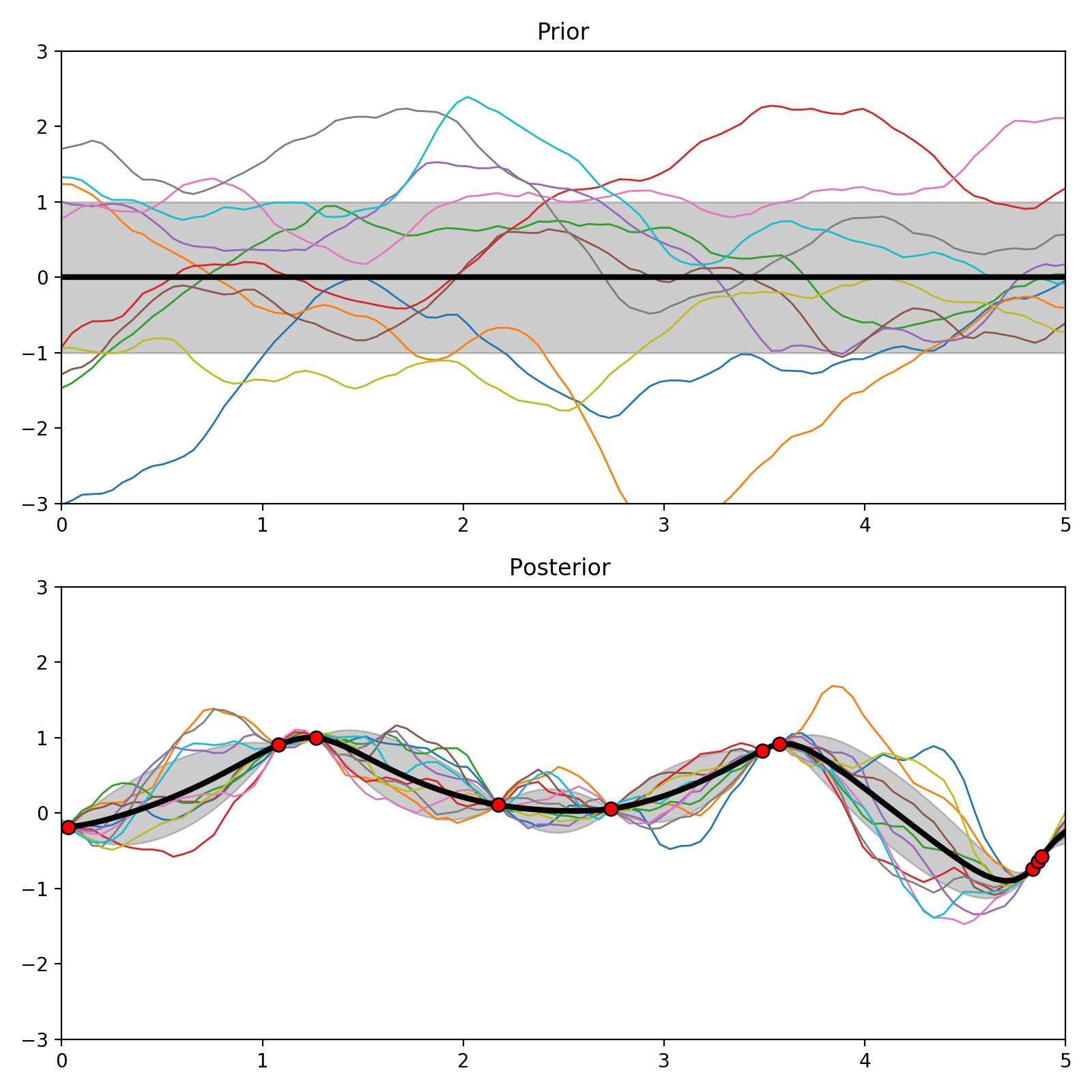}
\caption{\label{fig:matern_samp} Top: Prior Mean functions.
Bottom: Posterior mean functions.}
\end{subfigure}
\end{center}
\caption{Two Gaussian Processes fitted to 10 samples. Left: GP model with Squared Exponential Kernel, Right: GP model with Mat\`{e}rn.}
\end{figure}

In order to discuss the practical difference between the Mat\`{e}rn 3/2 kernel and the Squared Exponential, Figures~\ref{fig:sqrexp_samp} 
and~\ref{fig:matern_samp} are generated utilizing functions from the Scikit Learn framework~\cite{scikit-learn}. These figures contain
prior and posterior mean functions of the Gaussian Processes generated using the aforementioned covariance kernels. 
The prior mean functions sampled from the 
Gaussian Processes offer illustrations of typical functions that "live" in the function spaces that the covariance kernels expect. 
The sampled response variable follows the formula
$Y = \sin\left((X - 2.5)^2\right)$,
with 10 independent variable samples that follow $X\sim\mathcal{U}(0,5)$. 
Figure~\ref{fig:sqrexp_samp} contains the prior and posterior mean functions generated from Gaussian Process with the SE
Kernel using these response and independent variables. 
The gray space represents the uncertainty of the Gaussian Process models. 
For Figure~\ref{fig:matern_samp} the above process is repeated utilizing the Mat\`{e}rn 3/2 kernel instead of SE. 
Note that in Figure~\ref{fig:sqrexp_samp}, the prior and posterior mean functions are much smoother than the functions sampled from and 
produced by the GP with the Mat\'{e}rn kernel, as represented in Figure~\ref{fig:matern_samp}. 

\subsection{Maximum Likelihood Estimate for the Prior Mean}
\label{subsec:MLE}
The prior mean function that will be used is the \textit{maximum likelihood estimate for the prior 
mean}, calculated over the $5\times5$ square patch of pixels. 
This is done to change the character of the upsampling model so the model predicts the variation about the mean 
intensity in each sample.  
Typically, non-zero mean functions are used when there is an observed or assumed trend in the data. In the case of document images, pixel 
data is expected to retain certain intensities when inside a character or in the white space of a document. Because of these characteristics, a 
constant non-zero mean is chosen. Note that the derived prior mean functions is only constant over a single window, 
the prior mean will be constructed over each sample varies over the image. 

 To calculate the maximum likelihood estimate (MLE) for a constant prior mean, $\bar{\mathbf{f}} = f_0 \mathbf{1}$, 
 the Gaussian log-likelihood function is optimized with respect to $f_0$. 
 The log-likelihood is 
\begin{equation}
\ln \mathcal{L} = -\frac{1}{2}\left(\mathbf{f} - \bar{\mathbf{f}}\right)^T\mathbf{K}^T\left(\mathbf{f} - \bar{\mathbf{f}}\right)
- \frac{1}{2}\ln(\det |\mathbf{K}|) - \frac{N}{2}\ln(2\pi).
\label{eq:loglike}
\end{equation}
The maximum is calculated by setting the derivative of Equation~\ref{eq:loglike} with respect to $f_0$ and solving for $f_0$. 
Therefore the maximum likelihood estimate for the prior mean is: 
\begin{equation}
f_0 = \frac{\mathbf{1}^T\mathbf{K}^{-1}\mathbf{f}}{\mathbf{1}^T\mathbf{K}^{-1}\mathbf{1}}.
\label{eq:mle}
\end{equation}
Also, this maximum likelihood estimate for the prior mean can be recast as 
\[ f_0 = \frac{\left(\sum_i \mathbf{K}^{-1}_{[i]}\right) \cdot \mathbf{f} }{\sum_{i,j} \mathbf{K}^{-1}_{[i,j]}}. \] 
This interpretation is simply a weighted average with respect to the GP model. 

\subsection{Algorithm}
\label{subsec:OCRalgo}
In this upsampling algorithm, single channel grayscale document images are used. 
The Gaussian Process upsampling algorithm begins with the construction of the model weights with a length scale parameter derived from 
the original resolution of the image - $\ell = 20\min(1/h, 1/w)$. The upsampling ratio dictates the number of weight vectors needed, 
for example, when upsampling $4\times$, 16 new pixels are generated and therefore 16 weight vectors are needed. These vectors are
generated by utilizing the Cholesky factorization of $\mathbf{K}$ and then applying back substitution to calculate each 
$\mathbf{k}_*^T\mathbf{K}^{-1}$. The key factor is that the covariance kernel utilized in this methodology is isotropic-- it only depends on 
the distance between samples. Since a sliding window is used, the upsampling weights only need to be calculated once and can be used 
throughout the image. This is because the distance between sample pixels are related to their pixel index $(i,j)$ and the distance between each of 
the upsampled pixels and the rest of the window is identical for every window. 

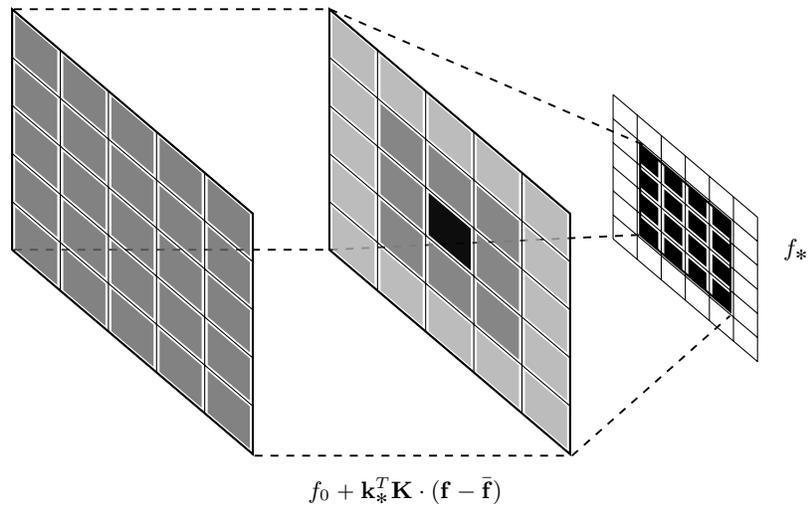
\begin{figure}
\begin{center}
 \begin{tikzpicture}[scale=.8,every node/.style={minimum size=1cm}]
 \draw[line width=0.25mm, black, dashed](-7, 4) -- (-1.75,4);
\draw[line width=0.25mm, black, dashed](-7, 0) -- (-1.75, 0);
\draw[line width=0.25mm, black, dashed](-3, -3.425) -- (2.275, -3.425);	
\draw[line width=0.25mm, black, dashed](-1.75, -0) -- (3.45,.25);
\draw[line width=0.25mm, black, dashed](2.275, -3.425) -- (4.9,-1.1);
\draw[line width=0.25mm, black, dashed](-1.75,4) -- (3.45,1.75);
\draw (-0.5, -4) node{ $f_0 + \mathbf{k}_*^T\mathbf{K}\cdot(\mathbf{f} - \bar{\mathbf{f}})$};
\draw (6, 0) node{$f_*$};

\begin{scope}[xshift=-200, yshift=0,every node/.append style={yslant=0.85,xslant=0}, yslant=-0.85,xslant=0] 
\fill[white,fill opacity=0.25] (0.,0.) rectangle (4,4); 
\draw[step=8mm, black] (0,0) grid (4,4); 
\draw[black,thick] (0,0) rectangle (4,4);
\fill[gray!70!black, fill opacity=0.75] (3.25, 3.25) rectangle (3.95, 3.95);
\fill[gray!70!black, fill opacity=0.75] (3.25, 2.45) rectangle (3.95, 3.15);
\fill[gray!70!black, fill opacity=0.75] (3.25, 1.65) rectangle (3.95, 2.35);
\fill[gray!70!black, fill opacity=0.75] (3.25, 0.85) rectangle (3.95, 1.55);
\fill[gray!70!black, fill opacity=0.75] (3.25, 0.05) rectangle (3.95, 0.75);

\fill[gray!70!black, fill opacity=0.75] (2.45, 0.05) rectangle (3.15, 0.75);
\fill[gray!70!black, fill opacity=0.75] (2.45, 3.25) rectangle (3.15, 3.95);

\fill[gray!70!black, fill opacity=0.75] (1.65, 0.05) rectangle (2.35, 0.75);
\fill[gray!70!black, fill opacity=0.75] (1.65, 3.25) rectangle (2.35, 3.95);

\fill[gray!70!black, fill opacity=0.75] (0.85, 0.05) rectangle (1.55, 0.75);
\fill[gray!70!black, fill opacity=0.75] (0.85, 3.25) rectangle (1.55, 3.95);

\fill[gray!70!black, fill opacity=0.75] (0.05, 3.25) rectangle (0.75, 3.95);
\fill[gray!70!black, fill opacity=0.75] (0.05, 2.45) rectangle (0.75, 3.15);
\fill[gray!70!black, fill opacity=0.75] (0.05, 1.65) rectangle (0.75, 2.35);
\fill[gray!70!black, fill opacity=0.75] (0.05, 0.85) rectangle (0.75, 1.55);
\fill[gray!70!black, fill opacity=0.75] (0.05, 0.05) rectangle (0.75, 0.75);

\fill[gray!70!black, fill opacity=0.75] (1.65,2.45) rectangle (2.35,3.15); 
\fill[gray!70!black, fill opacity=0.75] (3.15,2.35) rectangle (2.45,1.65); 
\fill[gray!70!black, fill opacity=0.75] (1.65,2.35) rectangle (2.35,1.65); 
\fill[gray!70!black, fill opacity=0.75] (0.85,2.35) rectangle (1.55,1.65); 
\fill[gray!70!black, fill opacity=0.75] (1.65,1.55) rectangle (2.35,0.85); 
\fill[gray!70!black, fill opacity=0.75] (0.85,0.85) rectangle (1.55, 1.55);
\fill[gray!70!black, fill opacity=0.75] (3.15,0.85) rectangle (2.45, 1.55);
\fill[gray!70!black, fill opacity=0.75] (3.15,2.45) rectangle (2.45, 3.15);
\fill[gray!70!black, fill opacity=0.75] (0.85, 2.45) rectangle (1.55, 3.15);
\end{scope}

\begin{scope}[xshift=-50, yshift=0,every node/.append style={yslant=0.85,xslant=0}, yslant=-0.85,xslant=0 ] 
\fill[white,fill opacity=0.25] (0.,0.) rectangle (4,4); 
\draw[step=8mm, black] (0,0) grid (4,4); 
\draw[black,thick] (0,0) rectangle (4,4);
\fill[gray!70!white, fill opacity=0.75] (3.25, 3.25) rectangle (3.95, 3.95);
\fill[gray!70!white, fill opacity=0.75] (3.25, 2.45) rectangle (3.95, 3.15);
\fill[gray!70!white, fill opacity=0.75] (3.25, 1.65) rectangle (3.95, 2.35);
\fill[gray!70!white, fill opacity=0.75] (3.25, 0.85) rectangle (3.95, 1.55);
\fill[gray!70!white, fill opacity=0.75] (3.25, 0.05) rectangle (3.95, 0.75);

\fill[gray!70!white, fill opacity=0.75] (2.45, 0.05) rectangle (3.15, 0.75);
\fill[gray!70!white, fill opacity=0.75] (2.45, 3.25) rectangle (3.15, 3.95);

\fill[gray!70!white, fill opacity=0.75] (1.65, 0.05) rectangle (2.35, 0.75);
\fill[gray!70!white, fill opacity=0.75] (1.65, 3.25) rectangle (2.35, 3.95);

\fill[gray!70!white, fill opacity=0.75] (0.85, 0.05) rectangle (1.55, 0.75);
\fill[gray!70!white, fill opacity=0.75] (0.85, 3.25) rectangle (1.55, 3.95);

\fill[gray!70!white, fill opacity=0.75] (0.05, 3.25) rectangle (0.75, 3.95);
\fill[gray!70!white, fill opacity=0.75] (0.05, 2.45) rectangle (0.75, 3.15);
\fill[gray!70!white, fill opacity=0.75] (0.05, 1.65) rectangle (0.75, 2.35);
\fill[gray!70!white, fill opacity=0.75] (0.05, 0.85) rectangle (0.75, 1.55);
\fill[gray!70!white, fill opacity=0.75] (0.05, 0.05) rectangle (0.75, 0.75);

\fill[gray,fill opacity=0.95] (1.65,2.45) rectangle (2.35,3.15); 
\fill[gray,fill opacity=0.95] (3.15,2.35) rectangle (2.45,1.65); 
\fill[black,fill opacity=0.95] (1.65,2.35) rectangle (2.35,1.65); 
\fill[gray,fill opacity=0.95] (0.85,2.35) rectangle (1.55,1.65); 
\fill[gray,fill opacity=0.95] (1.65,1.55) rectangle (2.35,0.85); 
\fill[gray,fill opacity=0.95] (0.85,0.85) rectangle (1.55, 1.55);
\fill[gray,fill opacity=0.95] (3.15,0.85) rectangle (2.45, 1.55);
\fill[gray,fill opacity=0.95] (3.15,2.45) rectangle (2.45, 3.15);
\fill[gray,fill opacity=0.95] (0.85, 2.45) rectangle (1.55, 3.15);
\end{scope}

 \begin{scope}[ xshift =50, yshift=0,every node/.append style={ yslant=0.85,xslant=0},yslant=-0.85,xslant=0 ] 
\draw[step=4mm, black] (1.2,1.2) grid (3.6,3.6); 
\draw[black,thick] (1.65,1.65) rectangle (3.15,3.15);
\fill[black] (2.05,2.05) rectangle (2.35,2.35); 
\fill[black] (1.65,2.05) rectangle (1.95,2.35); 
\fill[black] (2.45,2.05) rectangle (2.75,2.35); 
\fill[black] (2.05,2.45) rectangle (2.35,2.75); 
\fill[black] (2.05,1.95) rectangle (2.35,1.65); 
\fill[black] (1.65,2.45) rectangle (1.95,2.75); 
\fill[black] (2.45,2.45) rectangle (2.75,2.75); 
\fill[black] (2.75,1.95) rectangle (2.45,1.65); 
\fill[black] (1.65,1.95) rectangle (1.95,1.65); 
\fill[black] (3.15,1.95) rectangle (2.85,1.65);
\fill[black] (3.15,2.35) rectangle (2.85,2.05);
\fill[black] (3.15,2.75) rectangle (2.85,2.45);
\fill[black] (3.15,3.15) rectangle (2.85,2.85);
\fill[black] (2.75,3.15) rectangle (2.45,2.85);
\fill[black] (2.35,3.15) rectangle (2.05,2.85);
\fill[black] (1.95,3.15) rectangle (1.65,2.85);
\end{scope} %
\end{tikzpicture}
\end{center}
\caption{\label{fig:GP25} Schematic for the 5$\times$5 GP model for 4$\times$ upsampling. The completely gray grid illustrates the 
computation of $f_0$ over the sample, while the GP model combination on the second grid. 
The last grid illustrates the 16 new $f_*$ generated by combining the two, effectively replacing the pixel $(i,j)$}
\end{figure}

When performing  
upsampling over the document image, a sliding $5\times5$ pixel window is used as the sample for the GP model. Figure~\ref{fig:GP25} 
helps illustrate the sliding window GP method. The figure contains 3 grids of pixels. The first grid represents the constant maximum likelihood 
estimate for the prior mean over this pixel grid. The second grid represents the deviation of the sampled pixel values from the MLE. Together, 
these grids combine to interpolate 16 new pixels, replacing the pixel in the $(i,j)$ location. 

 In the implementation of this algorithm, the maximum likelihood estimate for the prior mean is generated when the 
$5\times5$ sample is loaded. Then each GP weight vector $\mathbf{k}_*^T\mathbf{K}^{-1}$ is applied to the residual between the 
MLE and pixels in the sampled window to model the deviation. The deviation and the MLE are combined to generate each new pixel $f_*$. 

\begin{figure}
\begin{center}
\begin{subfigure}{\textwidth}
\centering
\includegraphics[width=0.9\textwidth]{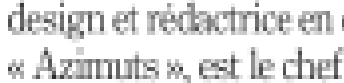}
\end{subfigure}
\begin{subfigure}{\textwidth}
\centering
\includegraphics[width=0.9\textwidth]{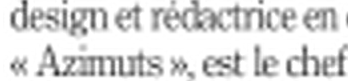}
\end{subfigure}
\caption{\label{fig:GPup25} Section of Page 154 of the LRDE dataset. Top: $4\times$ downsampled image crop.
 Bottom: $4\times$ GP upsampled. }
\end{center}
\end{figure}

As an example, Figure~\ref{fig:GPup25} is used to illustrate the upsampling results utilizing this Gaussian Processes algorithm. 
The top image in the figure is
the low resolution image (resized by copying the nearest pixels to be the same size as the GP image), 
and bottom text is from the GP upsampled image. When Tesseract is used on these images, it yields 
the following texts. The low resolution image Tesseract output is: 
\begin{quote}
"\textit{desigm \pounds r\'{e}dacimice en
fiflanEm, Et le chiet}",
\end{quote}
which is not an accurate representation of the ground truth. 
However, for the GP upsampled image, Tesseract generates 
\begin{quote}
"\textit{design et regactnce en
$<<$ Azzmuts $>>$. est le chef}".
\end{quote}
 It is clear that the GP upsampled version is much closer to the 
ground truth text of 
\begin{quote}
"\textit{design et r\'{e}dactrice en $<<$Azimuts $>>$, est le chef}".
\end{quote}

Tesseract works best when used on near-binary images as an input. In this case, near binary means that the majority of the pixels in the image 
are close to $0$ if they are within a character, or $255$ if not. However, sometimes the single channel images are calculated from RGB images 
that yield other shades of gray. In this case some images processing techniques can be used to better ``binarize'' these images. Aside from 
binarization, images can contain noise or textures within them, which can negatively effect the detection of characters. 
 A common way to handle excess noise and textures is to use a blurring operation to smooth out those regions. However, utilizing 
these blur convolutions can lead to unwanted removal of edges. 

To remove noise and textures without compromising edges the bilateral filtering approach illustrated by Tomasi and Manduchi is 
used~\cite{bilateral}. Bilateral filters 
reduce noise and textures without compromising edges, that is, without compromising the upsampled edges generated in the
 GP upsampling. 

If the image is not approximately binary, a thresholding technique can be used to force the text to be truly black. An adaptive Gaussian threshold 
process is used to generate binary images. 
Thresholding utilizes a set intensity value and replaces all pixels below that value to black and all pixels above the threshold to white. 
If there are shadows in the image, global thresholding can lead to large portions of the image to be blacked out. This could result in the 
majority of words in a document image to become inaccessible.  An adaptive-thresholding technique utilizes a neighborhood of pixels 
and calculates the threshold value locally to perform binarization. With Adaptive Gaussian thresholding, the threshold value is the 
weighted sum of neighborhood pixels in a Gaussian window~\cite{GaussWindow,opencv_library}. 

\begin{figure}
\begin{center}
\includegraphics[width=0.8\textwidth]{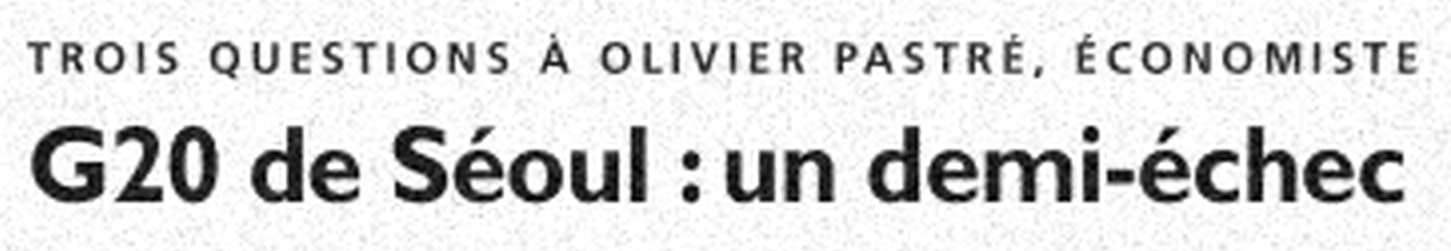}
\includegraphics[width=0.8\textwidth]{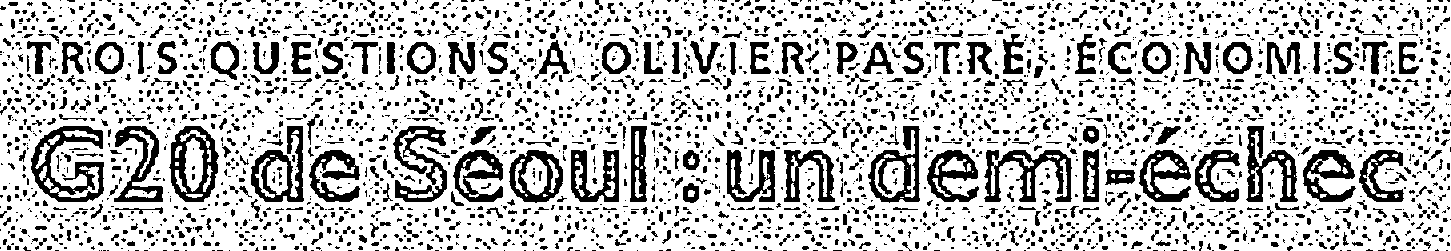}
\includegraphics[width=0.8\textwidth]{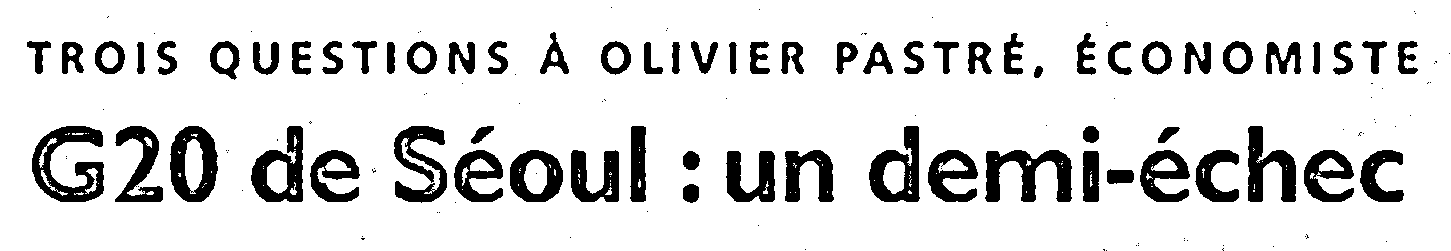}
\end{center}
\caption{\label{fig:thresh} Top: Noisy grayscale GP upsampled text block. Middle: Adaptive thresholding with no filter. Bottom: GP upsampled image with bilateral filter and adaptive thresholding.}
\end{figure}

Figure~\ref{fig:thresh} contains the results 
of the pipeline for processing low resolution images and is a visual explanation why filtering is necessary, especially when performing binarization. 
The top image is a GP upsampled version of a noisy low resolution image. 
The middle image is a thresholded version of the noisy image without using the bilateral image filter. 
Binarization, in this case, enhances the inherent noise, resulting in Tesseract to detect no characters. 
The bottom image is the noisy input image 
with bilateral filtering applied, and then thresholded. With the last image the Tesseract engine can detect every character. 

\begin{figure}
\begin{center}
\includegraphics[width=0.5\textwidth]{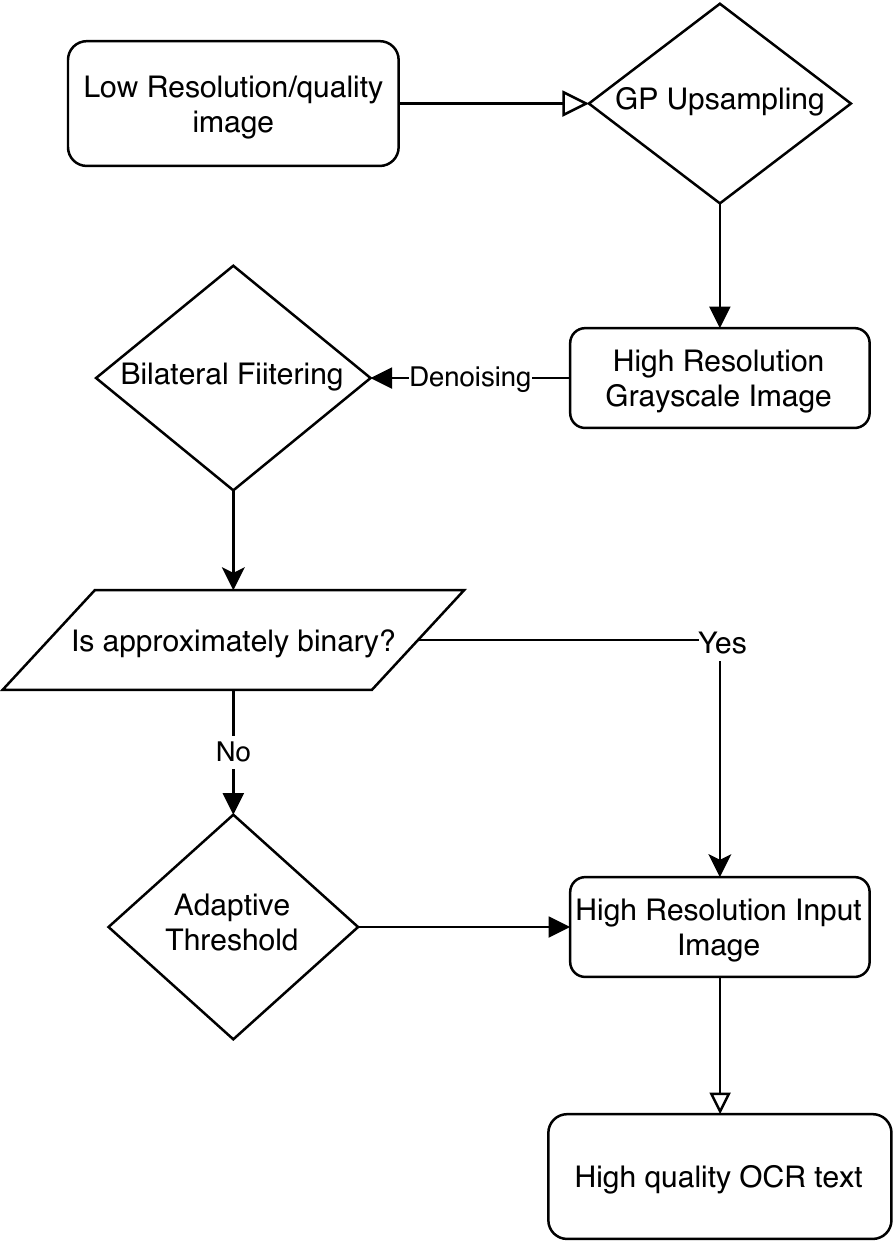}
\end{center}
\caption{\label{fig:OCRpipeline} The image processing pipeline used for higher quality OCR.} 
\end{figure}

The OCR pipeline used is as follows. 
First, a low resolution image is upsampled using the GP model presented earlier. 
Then, noise and unwanted textures from the high resolution image are removed 
while preserving edges by utilizing bilateral filtering. After the GP upsampled image is filtered, if the image is not approximately binary, 
an adaptive thresholding technique is 
used to convert the filtered high resolution image into a binary image to be ingested by the Tesseract OCR engine. 
For clarity, Figure~\ref{fig:OCRpipeline} contains an algorithmic diagram with each process.

\section{Testing}
In order to test the methodology, the EPITA Research and Development Laboratory (LRDE) dataset from~\cite{LRDE} is used. 
This dataset is publicly available but is
copyrighted, \textcopyright  2012 EPITA Research and Development Laboratory (LRDE) with permission from Le Nouvel Observateur. This 
dataset is based on the French
magazine Le Nouvel Observateur, issue 2402, November 18th-2th, 2010. The original images come from this magazine, and LRDE has 
generated the ground truth OCR 
from these images. This dataset is free for research, evaluation, and illustration and can be downloaded from 
\href{https://www.lrde.epita.fr/dload/olena/datasets/dbd/1.0/}
{LRDE's website}. 

To test the proposed GP upsampling algorithm, 
the original images' resolution is downsampled $4\times$ in width and height. Then these low resolution representations are 
 combined with Gaussian noise. Next, the noisy low resolution images are upsampled using the GP 
method illustrated in this manuscript. Finally, the upsampled images are then passed through the image processing pipeline illustrated in 
Figure~\ref{fig:OCRpipeline}, to extract detected characters. 

For this purpose, accuracy is calculated by comparing the number of words detected in the upsampled document to those that are present in the 
ground truth text. This is a fairly conservative measure, as increased accuracy in upsampling can lead to increased similarity in generated 
words with the true words. However, in this case, number of true words matched is a more direct measurement of accuracy that will effect 
applications that utilize image extracted text. 

\begin{figure}
\begin{center}
\includegraphics[width=\textwidth]{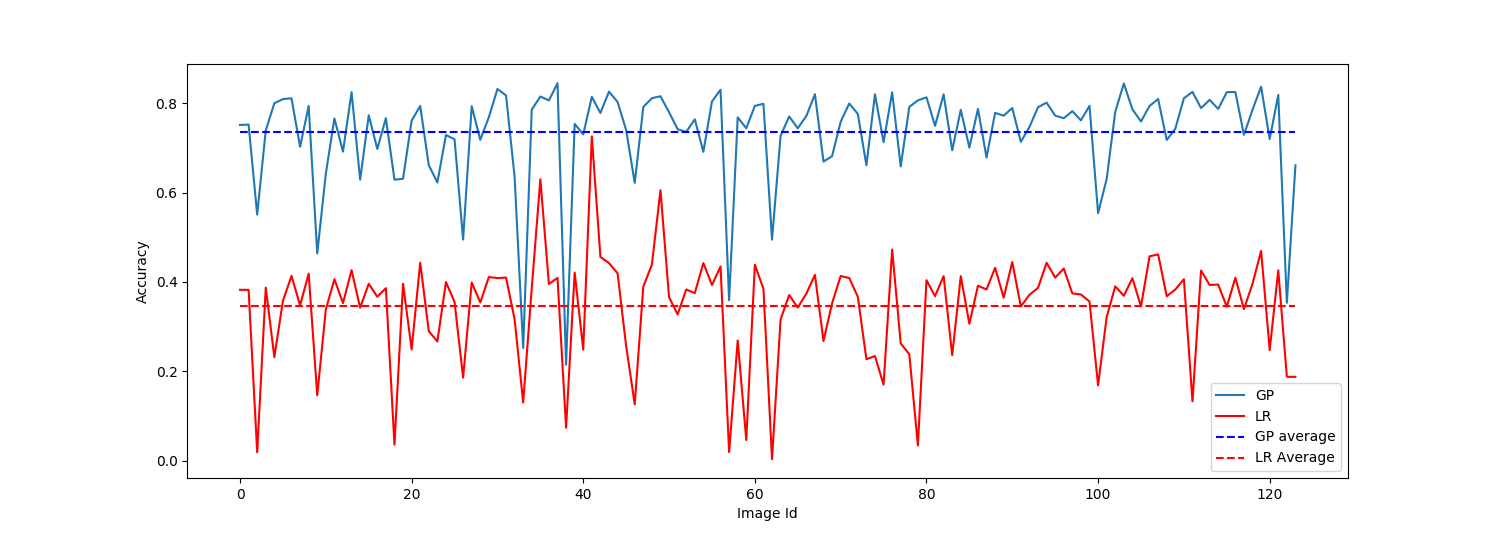}
\caption{\label{fig:acc1} GP upsampled OCR accuracy vs the Low Resolution accuracy with dashed lines denoting average accuracies.}
\end{center}
\end{figure}

First, the accuracy of the GP method is compared to the OCR extracted utilizing the low resolution images. Figure~\ref{fig:acc1} contains
a graph comparing the accuracy of OCR obtained from the GP upsampled images against OCR from the low resolution images, 
for each image in the dataset. 
In the figure, the blue line 
represents the OCR accuracy for each GP upsampled image, whereas the red line is the OCR accuracy of the low resolution 
images. Flat dashed lines are included to illustrate the mean accuracy of each set. 
 There are several dips in the graph where both the upsampled accuracy and the low resolution accuracy are very low, these 
pages of the magazine are comprised of mostly images where text is not the dominant feature. The extraneous information limits the 
capabilities of the Tesseract OCR engine. 

\begin{figure}
\begin{center}
\includegraphics[width=0.7\textwidth]{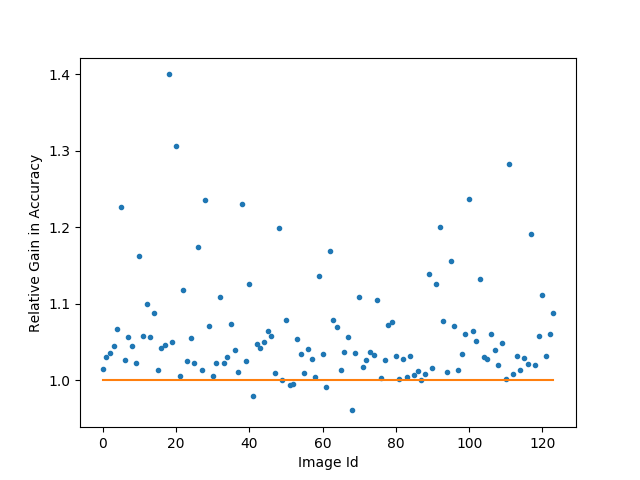}
\caption{\label{fig:ocr_compare} The relative gain utilizing GP upsampling vs bicubic over the noisy low resolution test set. The blue dots 
are the individual accuracy gains, and a reference line corresponding to equal accuracy is plotted in orange.}
\end{center}
\end{figure}

Most applications that require OCR will upsample sufficiently low resolution images. So, naturally, the GP algorithm is compared 
against the bicubic interpolation method, a common baseline in upsampling algorithms.  
For this implementation the bicubic method used is contained in the Python Image Library~\cite{bicubic}. 
In this test, the text generated by the 
GP based pipeline is compared against an analogous bicubic interpolation based pipeline. Figure~\ref{fig:ocr_compare} contains a 
plot of the relative gain in accuracy when utilizing GP over bicubic interpolation over the LRDE dataset. In the figure, the relative gain is 
depicted by the blue dots for each image in the dataset. Additionally, a line denoting equal performance is plotted as an orange line for
 reference.  For the majority of images, the proposed algorithm's extracted text better matches the ground truth text over the baseline interpolation.  
Some summary statistics are included in Table~\ref{tab:ocr}. The GP algorithm performs the best over the base low resoution images, and 
the bicubic interpolation based pipeline. The GP algorithm had the highest average accuracy, lowest variance and the highest minimum 
and maximum accuracy out of the three tests. The last column in the table is the relative gain in OCR accuracy by using the GP algorithm
instead of Bicubic or just using the low resolution image. There is a $6.26\%$ increase in character recognition against the 
bicubic upsampling. 

\begin{table}
  \begin{center}
    \caption{Summary statistics of the OCR accuracy over the LRDE subsampled dataset.}
    \label{tab:ocr}
    \begin{tabular}{l|c|c|c|c|c} 
       & Average & Variance & Max & Min & GP Relative Increase \\
      \hline
      GP & 0.735020 \hspace{1mm} & 0.012018  \hspace{1mm}& 0.844515  \hspace{1mm}& 0.214765  \hspace{1mm}&  N/A\\
      Bicubic & 0.695874 \hspace{1mm} & 0.013746  \hspace{1mm}& 0.835996  \hspace{1mm}& 0.175597  \hspace{1mm}& 6.26\% \\
      Low Resolution & 0.345170 \hspace{1mm} & 0.014018  \hspace{1mm}& 0.725663  \hspace{1mm}& 0.003584  \hspace{1mm}& 195\% \\
     \end{tabular}
  \end{center}
\end{table}

\section{Conclusion}
In this paper, a new Gaussian Process based interpolation model was produced for the explicit purpose of upsampling single-channel
 document images. Testing over a real-word data set revealed an increase in OCR accuracy over the baseline upsampling method,
  bicubic interpolation,  when used in conjunction with the Tesseract OCR engine. 

One could build a Gaussian Process model over the entire low resolution image and generate new pixels with inputs in a non-local sense. 
This provides issues in multiple areas. The kernel utilized in this context decay rapidly as distance is increased, so the new information 
gained will become less of a contribution than a hinderance when it comes to computation. Even though the weights are calculated 
using the Cholesky Factorization of the covariance matrix $\mathbf{K}$, the computational complexity of factorization is still $n^3/3$ 
where $n$ is the size of row and column size of $\mathbf{K}$~\cite{trefethen}. So even on a relatively small resolution image, say $500
\times500$, $\mathbf{K}$ will have size 250000, which will require $5.208\times10^{15}$ operations.  This is realistically infeasible, which 
leads well into the approach described in this paper. The windowed GP model can be reinterpreted as a Sparse Gaussian Processes 
that only utilizes information that is local to the interpolation pixels, which will have the most relevant information in both models. 

Some minor improvements could be gained by optimizing the length scale parameter, which could be found by maximizing the 
log-likelihood with respect to $\ell$. However, each window may have a different optimal length scale, which again, leads to an unwanted
increase in computational complexity. Additionally, one can tune $\ell$ for the dataset, but the value in this paper appears to be general, as it 
depends on the size of the low resolution image. 

Utilizing the proposed GP algorithm as an upsampling method for Optical Character Recognition yields on average a positive gain in 
accuracy versus a more traditional bicubic method when used to upsample the images for inputs to the Tesseract OCR engine. 
The GP algorithm uses a sliding window of $5\times5$ pixel sampled across the image. The yield in accuracy against bicubic can help text
based Natural Language Processing (NLP) models become perform better when placed in an end-to-end environment, like in financial 
applications, or for accessibility of documents and scanned images for people who are visually impaired.  

\section*{}
\label{sec:references}
\bibliographystyle{siam}
\bibliography{mybibfile}

\end{document}